\documentclass[acmlarge]{acmart}

\usepackage{booktabs} 
\usepackage{multirow}
\usepackage{algorithm}
\usepackage{algorithmic}

\acmDOI{0000001.0000001}

\begin{document}
\title{Cell Selection with Deep Reinforcement Learning in Sparse Mobile Crowdsensing}
\author{Leye Wang$^*$}
\affiliation{%
	\institution{Hong Kong University of Science and Technology}
	\country{Hong Kong, China}}
\author{Wenbin Liu$^*$}
\affiliation{%
	\institution{Jilin University}
	\country{China}
}
\author{Daqing Zhang}
\affiliation{%
	\institution{Peking University}
	\country{China}}
\author{Yasha Wang}
\affiliation{%
	\institution{Peking University}
	\country{China}}
\author{En Wang}
\affiliation{%
	\institution{Jilin University}
	\country{China}}
\author{Yongjian Yang}
\affiliation{%
	\institution{Jilin University}
	\country{China}
}

\begin{abstract}
Sparse Mobile CrowdSensing (MCS) is a novel MCS paradigm where data inference is incorporated into the MCS process for reducing sensing costs while its quality is guaranteed. Since the sensed data from different cells (sub-areas) of the target sensing area will probably lead to diverse levels of inference data quality, \textit{cell selection} (i.e., choose which cells of the target area to collect sensed data from participants) is a critical issue that will impact the total amount of data that requires to be collected (i.e., data collection costs) for ensuring a certain level of quality. To address this issue, this paper proposes a \textit{Deep Reinforcement} learning based \textit{Cell} selection mechanism for Sparse MCS, called \textit{DR-Cell}. First, we properly model the key concepts in reinforcement learning including \textit{state}, \textit{action}, and \textit{reward}, and then propose to use a deep recurrent Q-network for learning the Q-function that can help decide which cell is a better choice under a certain state during cell selection. Furthermore, we leverage the transfer learning techniques to reduce the amount of data required for training the Q-function if there are multiple correlated MCS tasks that need to be conducted in the same target area. Experiments on various real-life sensing datasets verify the effectiveness of DR-Cell over the state-of-the-art cell selection mechanisms in Sparse MCS by reducing up to 15\% of sensed cells with the same data inference quality guarantee.
\end{abstract}

%
%


%
%



\maketitle

\renewcommand{\shortauthors}{Wang et al.}

\section{Introduction}

With the prevalence of smart mobile devices, mobile crowdsensing (MCS) becomes a novel sensing mechanism nowadays to address various urban tasks such as environment and traffic monitoring~\cite{zhang20144w1h}. Traditional MCS mechanisms usually collect a large amount of data to cover almost all the \textit{cells} (i.e., subareas) of the target area to ensure quality. This requires MCS organizers to recruit many participants (e.g., at least one from every cell for full coverage), leading to a relatively high cost. To reduce such cost while still ensuring a high level of quality, a new MCS paradigm, namely \textit{Sparse MCS}, is proposed recently~\cite{Wang2017SPACE,wang2016sparse}.  Sparse MCS collects data from only a few cells while intelligently inferring the data of rest cells with quality guarantees (i.e., the error of inferred data is lower than a threshold). Hence, compared to traditional mechanisms, MCS organizers' cost can be reduced since only a few participants need to be recruited, while the task quality is still ensured.

\begin{figure}[t]
	\begin{center}
		\includegraphics[width=.8\linewidth] {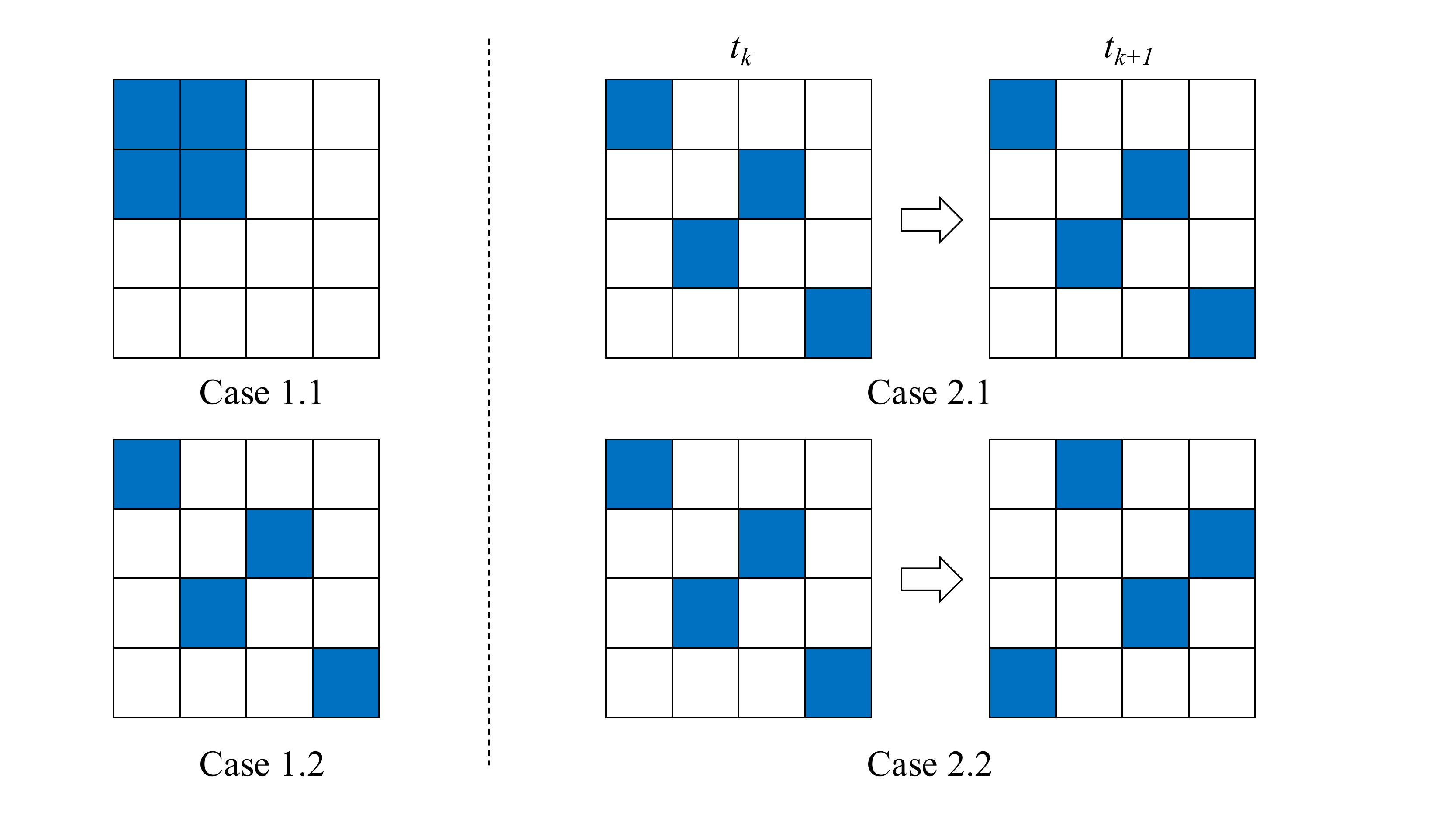}
		\vspace{-1em}
		\caption{Different cell selection cases.}
		\label{fig:cell_selection_case}
	\end{center}
\end{figure}

In Sparse MCS, one key issue affecting how much cost can be practically saved is \textit{cell selection} --- \textit{which cells the organizer decides to collect sensed data from participants}~\cite{wang2016sparse}. To show the importance of cell selection, Figure~\ref{fig:cell_selection_case} (left part) gives an illustrative example of two different cell selection cases in a city, which is split to $4\times 4$ cells. In Case 1.1, all the selected cells are gathered in one corner of the city; in Case 1.2, the collected data is more widely distributed in the whole city. As data of most sensing tasks has spatial correlations (i.e., nearby cells may have similar data), e.g., air quality \cite{Zheng2013U}, the cell selection of Case 1.2 will generate a higher inference quality of the inferred data than Case 1.1. Moreover, a MCS campaign usually lasts for a long time (i.e., sensing every one hour), so that not only spatial correlations, but also temporal correlations need to be carefully considered in cell selection. As shown in Figure~\ref{fig:cell_selection_case} (right part), sensing the same cells in continuous cycles (Case 2.1) may not be as efficient as sensing the different cells (Case 2.2) considering the inference quality. While data of different MCS applications may involve diverse spatio-temporal correlations, determining the proper cell selection strategies is a non-trivial task.

Existing works on Sparse MCS mainly leverage \textit{Query-By-Committee} (QBC) \cite{wang2015ccs,Wang2017SPACE} in cell selection. QBC first uses various inference algorithms to deduce the data of all the unsensed cells, and then chooses the cell where the inferred data of various algorithms have the largest variance as the next cell for sensing. Briefly, the cell selection criteria of QBC is choosing the cell which is the most uncertain considering a committee of inference algorithms, i.e., hard-to-infer. While QBC has shown its effectiveness in some scenarios \cite{wang2015ccs,Wang2017SPACE}, it does not directly optimize the objective of Sparse MCS, i.e., minimizing the number of sensed cells under a quality guarantee. In fact, the existing works using QBC also realize that its performance is still far from the optimal cell selection strategy\footnote{Please note that the optimal cell selection strategy is impractical as it needs to know the ground truth data of each cell in advance, which is absolutely impossible in reality~\cite{wang2015ccs}.} \cite{wang2015ccs}. To reduce this performance gap from the optimal strategy, our research question here is, \textit{can we find a better cell selection strategy in Sparse MCS, which can directly minimize the number of selected cells under the inference quality guarantee?}

To this end, in this paper, we design a new cell selection framework for Sparse MCS, called \textit{DR-Cell}, with \textit{Deep Reinforcement} learning. In recent years, deep reinforcement learning has shown its successes in decision making problems in diverse areas such as robot control \cite{gu2017deep} and game playing~\cite{silver2017mastering,mnih2015human}. In general, deep reinforcement learning can benefit a large set of decision making problems which can be abstracted as `an \textit{agent} needs to decide the \textit{action} under a certain \textit{state}'. Our cell selection problem can actually be interpreted as `\textit{an MCS server} (agent) needs to \textit{choose the next cell for sensing} (action) considering the \textit{data already collected} (state)'. In this regard, it is promising to apply deep reinforcement learning on the cell selection problem in Sparse MCS.

To effectively employ deep reinforcement learning in cell selection, we still face several issues. 

\begin{enumerate}
	\item The first issue is how to mathematically model the \textit{state}, \textit{action}, and \textit{reward}, which are key concepts in reinforcement learning \cite{Sutton2005Reinforcement}. Briefly speaking, reinforcement learning attempts to learn a \textit{Q-function} which takes the current \textit{state} as input, and generates \textit{reward} scores for each possible \textit{action} as output. Then, we can take the action with the highest reward score as our decision. Only if we can model state, action and reward properly, we can generate the cell selection policy that can minimize the number of cells selected under the quality requirement.
	\item The second issue is how to learn the \textit{Q-function}. Traditional Q-learning techniques in reinforcement learning work well in the scenarios where the state and action spaces are small (i.e., number of states and actions is limited). However, in Sparse MCS, the space of state is actually quite large. For example, suppose there are 100 cells (subareas) in the target sensing area, then even we only consider the current cycle, the possible number of states grows up to $2^{100}$ (whether a cell is sensed by participants or not). To overcome the difficulty of large state space, we hence propose to leverage deep learning along with reinforcement learning, i.e., deep reinforcement learning to learn Q-function for our cell selection problem.
	\item The last issue is the training data scarcity issue. Usually, deep reinforcement learning requires a lot of training data (i.e., known state, action, and reward) to learn  Q-function. In areas such as robot control or game playing, a robot or a computer can continuously run for data collection until the training performance is good. However, in MCS, we cannot have an unlimited amount of data for training. Then, how to address the training data scarcity issue, at least partially, should also be considered in our cell selection problem.
\end{enumerate}

In summary, this work has the following contributions:

(1) To the best of our knowledge, this work is the first research that attempts to leverage the deep reinforcement learning to address the critical question in Sparse MCS, cell selection.

(2) We propose \textit{DR-Cell} to select the best cell for obtaining the sensed data in Sparse MCS. More specifically, we model the \textit{state} with one-hot encoding, and the \textit{reward} following the inference quality requirement of Sparse MCS. Then, considering the spatio-temporal correlations hidden in the sensed data, we propose a recurrent deep neural network structure to learn the reward output from the inputs of state and action. Finally, to relieve the dependence on a large amount of training data, we propose a transfer learning algorithm between heterogeneous sensing tasks in the same target area, so that the decision function learned on one task can be efficiently transferred to another task with only a little training data.

(3) Experiments on real data of sensing tasks including temperature, humidity and air quality monitoring have verified the effectiveness of DR-Cell. In particular, DR-Cell can outperform the state-of-the-art mechanism QBC by reducing up to 15\% of cells while guaranteeing the same quality in Sparse MCS.
\section{Related Work}

\subsection{Sparse Mobile Crowdsensing}
MCS is proposed to utilize widespread crowds to perform large-scale sensing tasks \cite{zhang20144w1h,ganti2011mobile,guo2015mobile}. In practice, to minimize sensing cost while ensuring data quality, some MCS tasks involve inference algorithms to fill missing data of unsensed cells, such as noise sensing \cite{rana2010ear}, traffic monitoring \cite{zhu2013compressive}, and air quality sensing \cite{wang2015ccs}. It is worth noting that in such MCS tasks, compressive sensing \cite{candes2009exact,donoho2006compressed} has become the \textit{de facto} choice of the inference algorithm \cite{rana2010ear,zhu2013compressive,wang2015ccs,xu2015more,Wang2017SPACE}. Recently, by extracting the common research issues involved in such tasks involving data inference, Wang et al. \cite{wang2016sparse} propose a new MCS paradigm, called \textit{Sparse MCS}. Besides the inference algorithm, Sparse MCS also abstracts other critical research issues such as \textit{cell selection} and \textit{quality assessment}. Later, privacy protection mechanism is also added into Sparse MCS \cite{wang2016differential}. In this paper, we focus on the \textit{cell selection} issue and aim to use deep reinforcement learning techniques to address it.

\subsection{Deep Reinforcement Learning}

Reinforcement Learning (RL) \cite{Sutton2005Reinforcement} is concerned with how to map states to actions so as to maximize the cumulative rewards. It utilizes rewards to guide agent to do the better sequential decisions, and has substantive and fruitful interactions with other engineering and scientific disciplines. Recently, many researchers focus on combining deep learning with reinforcement learning to enhance RL in order to solve concrete problems in the sciences, business, and other areas. Mnih et al. \cite{Mnih2013Playing} propose the first deep reinforcement learning model (DQN) to deal with the high-dimensional sensory input successfully and apply it to play seven Atari 2600 games. More recently, Silver et al. \cite{Silver2016Mastering} apply DQN and present $AlphaGo$, which was the first program to defeat world-class players in Go. Moreover, to deal with the partially observable states, Hausknecht and Stone\cite{DRQN} introduce a deep recurrent neural network (DRQN), particularly a Long-Short-Term-Memory (LSTM) Network, and apply it to play Atari 2600 games. Lample and Chaplot \cite{Lample2016Playing} even use DRQN to play FPS Games.

While deep reinforcement learning has already been used in a variety of areas, like object recognition \cite{Ba2014Multiple}, robot control \cite{Levine2015End}, and communication protocol \cite{Foerster2016Learning}, MCS researchers just began to apply it very recently. Xiao et al. \cite{Xiao2017Mobile} formulate the interactions between a server and vehicles as a vehicular crowdsensing game. Then they propose the Q-learning based strategies to help server and vehicles make the optimal decisions for the dynamic game. Moreover, Xiao et al. \cite{Xiao2017A} apply DQN to derive the optimal policy for the Stackelberg game between a MCS server and a number of smartphone users. As far as we know, this paper is the first research attempt to use deep reinforcement learning in cell selection of sparse MCS,  so as to reduce MCS organizers' data collection costs while still guaranteeing the data quality.
\section{Problem Formulation}

We first define several key concepts, and then mathematically formulate the cell selection problem in Sparse MCS. Finally, we illustrate a running example to explain our problem in more details.

\textit{Definition 1. Sensing Area.} We suppose that the target sensing area can be split into a set of cells (e.g., $1km\times 1km$ grids~\cite{Zheng2013U,Wang2017SPACE}). The objective of a sensing task is to get a certain type of data (e.g., temperature, air quality) of all the cells in the target area.

\textit{Definition 2. Sensing Cycle.} We suppose the sensing tasks can be split into equal-length cycles, and the cycle length is determined by the MCS organizers according to their requirements \cite{xiong2015emc,Wang2017SPACE}. For example, if an organizer wants to update the data of the target sensing area every one hour, then he can set the cycle length to one hour.

\textit{Definition 3. Ground Truth Data Matrix.} Suppose we have $m$ cells and $n$ cycles, then for a certain sensing task, the ground truth data matrix is denoted $\mathcal D_{m\times n}$, where $\mathcal D[i,j]$ is the true data in cell $i$ at cycle $j$.

\textit{Definition 4. Cell Selection Matrix.} In Sparse MCS, we will only select partial cells in each cycle for data collection, while inferring the data for rest cells. Cell selection matrix, denoted $\mathcal S_{m \times n}$, marks the cell selection results. $\mathcal S[i,j]=1$ means that the cell $i$ is selected at cycle $j$ for data collection; otherwise, $\mathcal S[i,j]=0$.

\textit{Definition 5. Inferred Data Matrix.} In Sparse MCS, when an organizer decides not to collect any more data in the current cycle, the data of unsensed cells will then be inferred. Then, we denote the inferred data of the $k$-th cycle as $\hat{\mathcal D}[:,k]$, and thus the inferred data of all the cycles as a matrix $\hat{\mathcal D}_{m \times n}$. Note that in Sparse MCS, compressive sensing is the \textit{de facto} choice of the inference algorithm nowadays \cite{rana2010ear,zhu2013compressive,wang2015ccs,xu2015more,Wang2017SPACE}, and we also use it in this work.

\textit{Definition 6. ($\epsilon$, $p$)-quality}~\cite{Wang2017SPACE}. In Sparse MCS, the quality guarantee is called \textit{($\epsilon$, $p$)-quality}, meaning that in $p\cdot 100\%$ of cycles, the inference error (e.g., mean absolute error) is not larger than $\epsilon$. Formally,
\begin{equation}
	|\{k|error(\mathcal D[:,k], \hat{\mathcal D}[:,k]) \leq \epsilon, 1\leq k \leq n\}| \geq n \cdot p
\end{equation}
where $n$ is the number of total sensing cycles.

Note that in practice, since we do not know the ground truth data matrix $\mathcal D$, we also cannot know whether $error(\mathcal D[:,k], \hat{\mathcal D}[:,k])$ is smaller than $\epsilon$ in the current cycle with 100\% confidence. This is why we include $p$ in the quality requirement, as it is impossible to ensure 100\% of cycles' error less than $\epsilon$. To ensure ($\epsilon$, p)-quality, certain quality assessment method is needed in Sparse MCS to estimate the the probability of the error less than $\epsilon$ for the current cycle. If the estimated probability is larger than $p$, then the current cycle satisfies ($\epsilon$, p)-quality and no more data will be collected (we will then move to the next sensing cycle). In Sparse MCS, leave-one-out based Bayesian inference method is often leveraged for quality assessment \cite{Wang2017SPACE,wang2015ccs,wang2016sparse}, and we also use it in this work.

\textbf{Problem [Cell Selection]: Given a Sparse MCS task with $m$ cells and $n$ cycles, using compressive sensing as data inference method and leave-one-out based Bayesian inference as quality assessment method, we aim to select a minimal subset of sensing cells during the whole sensing process (minimize the number of non-zero entries in the \textit{cell-selection matrix} $\mathcal S$), while satisfying $(\epsilon,p)$-quality:}
\begin{align*}
	\vspace{-0.4em}
	&\min \,\, \sum_{i=1}^m \sum_{j=1}^n \mathcal S[i,j] \\
	&s.t., \textit{satisfy } \textit{($\epsilon$, $p$)-quality}
\end{align*}

\begin{figure}[t]
	\begin{center}
		\includegraphics[width=.9\linewidth] {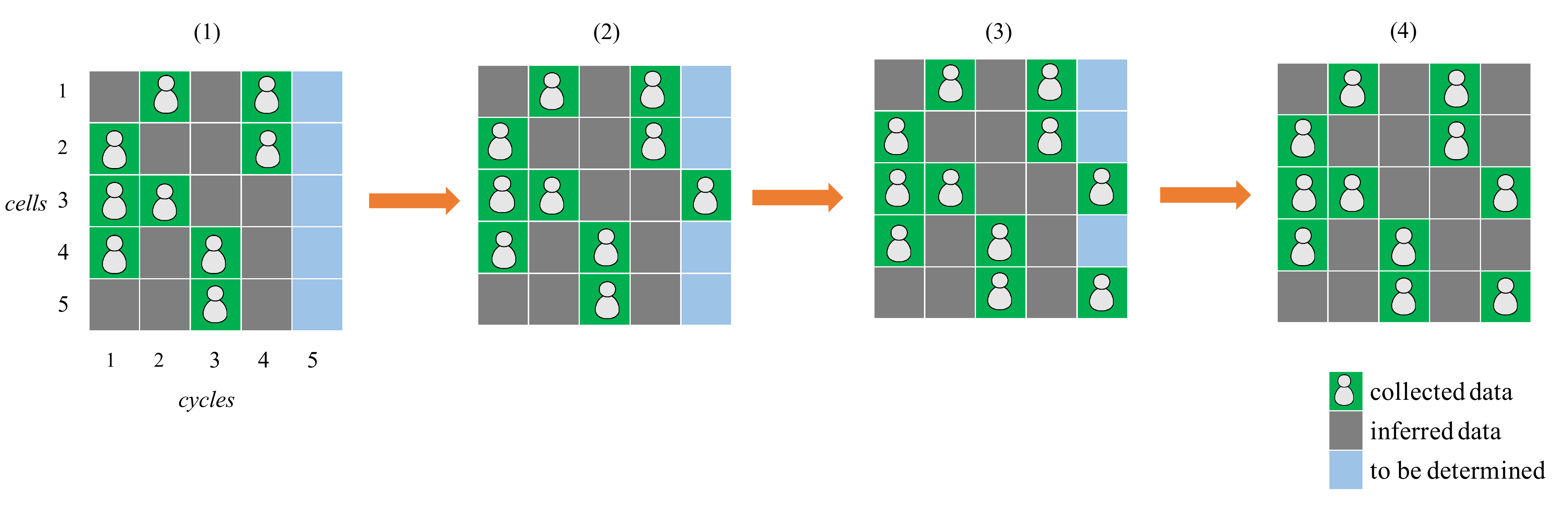}
		\vspace{-1em}
		\caption{Running example.}
		\label{fig:example}
	\end{center}
\end{figure}

We now use a running example to illustrate our problem in more details, as shown in Figure~\ref{fig:example}. (1) Suppose we have five cells and the current is the 5th cycle; (2) We select the cell 3 for collecting data, and then assess whether the current cycle can satisfy $(\epsilon, p)$-quality; (3) As we find that the quality requirement is not satisfied, we continue collecting data from cell 5; (4) The quality requirement is now satisfied, so the data collection is terminated for the current cycle, and the data of the unsensed cells is inferred. In this example, we see that after five cycles, there are totally 11 data submissions from participants. The objective of our cell selection problem is exactly to minimize the number of data submissions.

\section{Methodology}

\begin{figure}[t]
	\begin{center}
		\includegraphics[width=.6\linewidth] {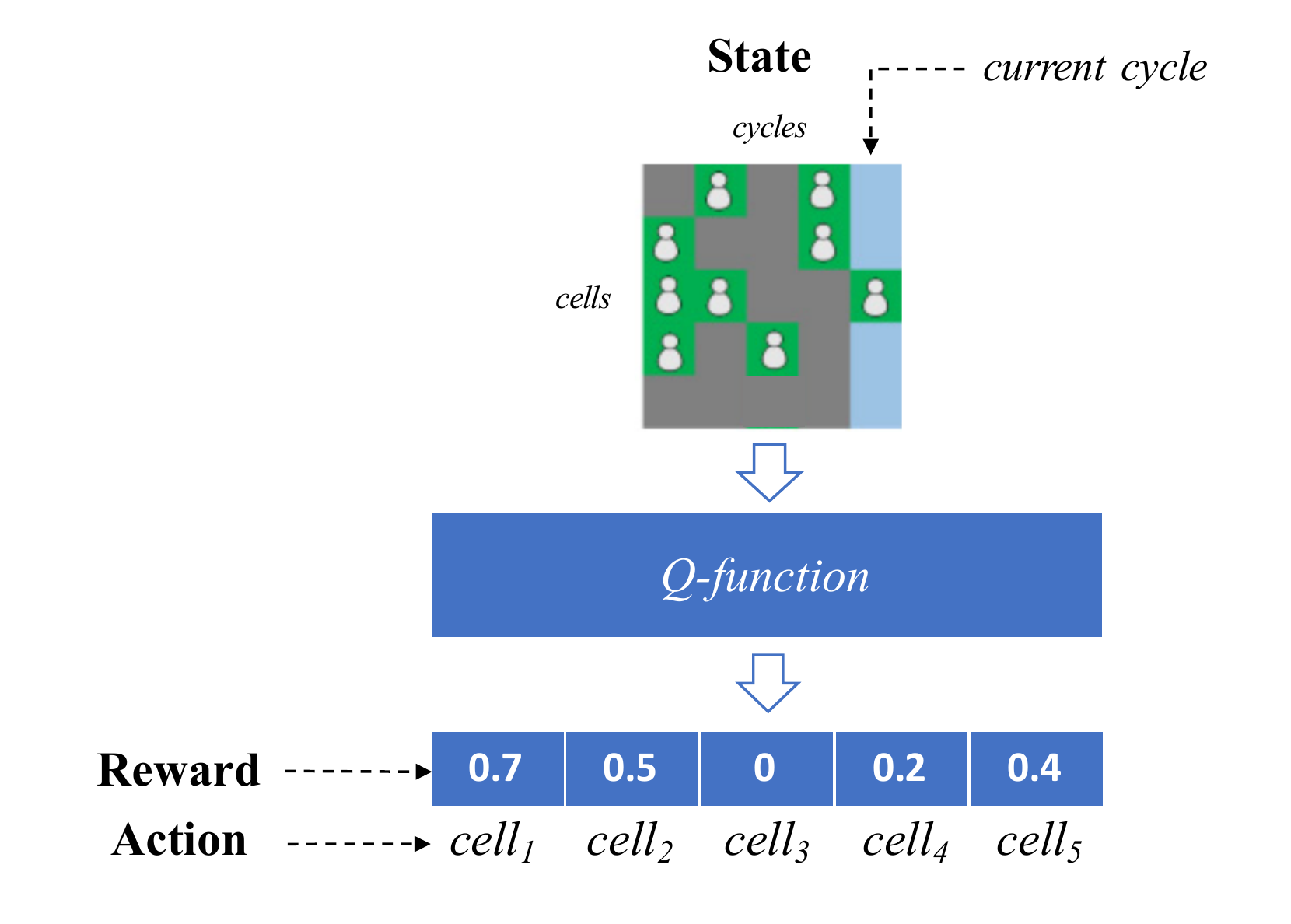}
		\vspace{-1em}
		\caption{State, action, reward in DR-CELL}
		\label{fig:framework}
	\end{center}
\end{figure}

In this section, we propose a novel mechanism, called \textit{DR-Cell}, to address the cell selection problem with deep reinforcement learning. First, we will mathematically model the state, reward, and action used in DR-Cell. Then, with a simplified MCS task example (i.e., there are only a few cells in the target area), we explain how traditional reinforcement learning works find the most appropriate cell for sensing based on our state, reward, and action modeling. Afterward, we elaborate how deep learning can be combined with reinforcement learning (i.e., deep reinforcement learning) to work on more realistic cases of cell selection where the target area can include a large number of cells. Finally, we describe how transfer learning can help us to generate a cell selection strategy with only a little training data under some specific conditions.


\subsection{Modeling state, action, and reward}

To apply deep reinforcement learning on cell selection, we need to model the key concepts in terms of state, action, and reward. Figure~\ref{fig:framework} illustrates the relationship between the three key concepts in DR-Cell. Briefly speaking, in DR-CELL, based on the current data collection \textit{state}, we need to learn a \textit{Q-function} (will be elaborated in next a few subsections), which can output \textit{reward} scores for each possible \textit{action}. The \textit{action} in cell selection is choosing \textit{which} cell as the next sensing cell, while \textit{reward} indicates how good a certain action is. If an action (i.e., a cell) gets a higher reward score, it may be a better choice. Next we formally model the three concepts.

\begin{figure}[t]
	\begin{center}
		\includegraphics[width=.7\linewidth] {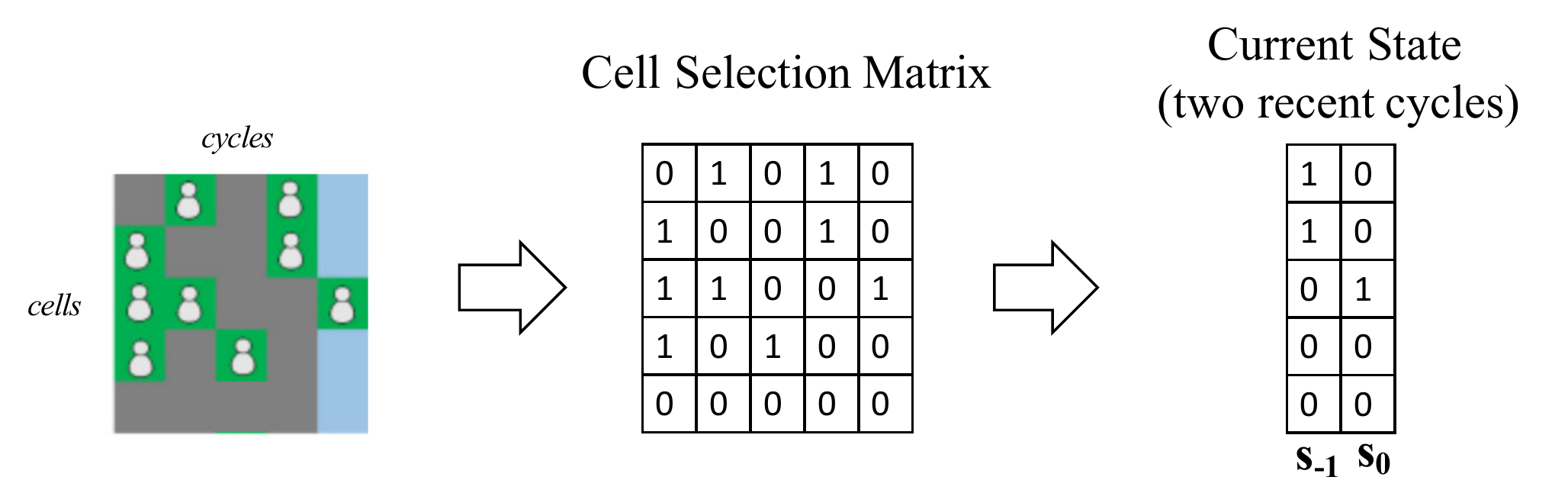}
		\vspace{-1em}
		\caption{An example of state model.}
		\label{fig:state}
	\end{center}
\end{figure}

(1) \textbf{State} represents the current data collection condition of the MCS task. In Sparse MCS, \textit{cell selection matrix} (Definition 4) can naturally model the state of Sparse MCS well, as it records both where and when we have collected data from the target sensing area during the whole task. In practice, we can just keep the recent $k$ cycles' cell selection matrix as the state, denoted as $\mathbf S=[\mathbf{s}_{-k+1},..., \mathbf{s}_{-1}, \mathbf{s}_0]$, where $\mathbf{s}_0$ represents the cell selection vector of the current cycle (1 means selected and 0 means no), $\mathbf{s}_{-1}$ represents last cycle, and so on. Figure~\ref{fig:state} shows an example of how we encode the current data collection condition into the state model if recent two cycles are considered. Note that we use $\mathbb S$ to denote the whole set of states. As an example, suppose that we consider the recent two cycles and there are totally five cells in the target area, then the number of possible states, i.e., $|\mathbb S| = 2^{2\times 5}=1024$.

(2) \textbf{Action} means all the possible decisions that we may make in cell selection. Suppose there are totally $m$ cells in the target sensing area, then our next selected cell can have $m$ choices, leading to the whole action set $\mathbb A = \{1, 2, \cdots, m \}$. Note that while in practice we will not select one cell for more than once in one cycle, to make the action set consistent under different states, we assume that the possible action set is always the complete set of all the cells under any state. More specifically, if some cells have already been selected in the current cycle, then the probability of choosing these cells is zero.

(3) \textbf{Reward} is used to indicate how good an action is. In each sensing cycle, we select actions one by one until the selected cells can satisfy the quality requirement in the current cycle (i.e., inference error less than $\epsilon$\footnote{When running Sparse MCS, we have to set a probability $p$ in quality requirement, i.e., ($\epsilon,p$)-quality, as we do not know the ground truth data of unsensed cells. However, in the training stage of the cell selection policy, we assume that we have obtained the data of all the cells in the target area for some time (e.g., 1 day), and thus we can directly compute the inference error. More details on the training stage will be described in the evaluation section.}). Satisfying this quality requirement is the goal of cell selection and should be reflected in the reward modeling. Hence, a positive reward, denoted by $R$, would be given to an action (i.e., a cell) under a state $\mathbf S$ if the quality requirement is satisfied in the current cycle after the action is taken. In addition, as selecting participants to collect data incurs cost, we also put a negative score $-c$ in the reward modeling of an action. Then, the reward can be written as $\mathbf{R}=q\cdot R-c$, in which $q \in \{0,1\}$ means whether the action makes the current cycle satisfy the inference quality requirement.




With the above modeling, we then need to learn the \textit{Q-function} (see Figure~\ref{fig:framework}) which can output the reward score of every possible action under a certain state. In the next subsection, we will first use a traditional reinforcement learning method, tabular Q-learning, to illustrate a simplified case where a small number of cells exist in the target sensing area.

\begin{algorithm}[t]
\renewcommand{\algorithmicrequire}{\textbf{Initialization:}}
\caption{Tabular Q-learning} 
\label{Alg1}
\begin{algorithmic}[1]
\REQUIRE ~~\\
    Q-table: $Q[\mathbf S,\mathbf A]=0,\ \forall \mathbf{S} \in \mathbb S,\ \forall \mathbf{A} \in \mathbb A$

\WHILE{True}

    \STATE{Update the current state $\mathbf{S}=[\mathbf{s}_{-k},..., \mathbf{s}_{-1}, \mathbf{s}_0]$}
    \STATE{Check Q-table, select and perform the action $A$, which has the largest Q-value via the $\delta$-greedy algorithm.}

    \IF{The selected cells in current cycle satisfy the quality requirements}
        \STATE{// \emph{The cell selection in this cycle is complete, next state is the initial state in next cycle.}}
        \STATE{$\mathbf{s}_1 = \mathbf{0}_{m\times 1}$}
        \STATE{Update the next state $\mathbf{S}'=[\mathbf{s}_{-k+1},..., \mathbf{s}_{0}, \mathbf{s}_1]$}
        \STATE{Obtain the reward for this action $\mathbf{R}=R-c$}
    \ELSE
        \STATE{// \emph{Continue to select cells in this cycle.}}
        \STATE{$\mathbf{s}_0'= \mathbf{s}_0 + [0,\cdots,0,1,0,\cdots,0]^T$ ($1$ is in the $\mathbf A$-th element)}
        \STATE{Update the next state $\mathbf{S}'=[\mathbf{s}_{-k},..., \mathbf{s}_{-1}, \mathbf{s}_0']$}. 
        \STATE{Obtain the reward for this action $\mathbf{R}=-c$}
    \ENDIF
    \STATE{Update Q-table via (\ref{eq1}) and (\ref{eq2}).}
\ENDWHILE
\end{algorithmic}
\end{algorithm}

\subsection{Training Q-function with Tabular Q-Learning}

In traditional reinforcement learning, a widely used strategy to obtain the Q-function is the tabular Q-learning. In this method, the Q-function is represented by a Q-table, denoted as $Q_{|\mathbb S|\times |\mathbb A|}$. Each element in the Q-table, $Q[\mathbf S,\mathbf A]$ represents the reward score of a certain action $\mathbf A \in \mathbb A$ under a certain state $\mathbf S \in \mathbb S$. The objective of learning the Q-function is then equivalent to filling all the elements in the Q-table.

The tabular Q-learning algorithm is shown in Algorithm 1. Under the current state $\mathbf{S}$, the algorithm selects the action which has the maximum value from $Q[\mathbf{S},\mathbf{A}],\ \forall \ \mathbf{A} \in \mathbb A$ (in fact, not always the best action is selected, will be elaborated later). After the action has been conducted, $i.e.$ the cell has been selected and the data of the cell has been collected, the current state will change to the next state $\mathbf{S}'$. Note that if the current cycle satisfies the quality requirement (i.e., inference error less than $\epsilon$), then the next state will shift to a new cycle. For the selected action, we would get the real reward $\mathbf{R}$ considering whether the inference quality requirement of the current cycle is satisfied and then update Q-table according to the equations as follows
{\begin{align}
Q[\mathbf{S},\mathbf{A}]=(1 - \alpha)Q[\mathbf{S},\mathbf{A}]+\alpha \big( \mathbf{R} +\gamma V(\mathbf{S}') \big),
\label{eq1}
\end{align}}%
{\begin{align}
V(\mathbf{S}')=max_{\mathbf{A}'} Q[\mathbf{S}',\mathbf{A}'], \forall \mathbf{A}' \in \mathbb A
\label{eq2}
\end{align}}%
where $V(\mathbf{S}')$ provides the highest expected reward score of the next state $\mathbf{S}'$ (i.e., the reward of the best action under the next state $\mathbf S'$); $\gamma\in[0,1]$ is the discount factor indicating the myopic view of the Q-learning regarding the future reward; $\alpha\in(0,1]$ is the learning rate.

Moreover, during the training stage, under a certain state, if we always select the action with the largest reward score in the Q-table, the algorithm may get a local optima. To address this issue, we need to \textit{explore} during training, i.e., sometimes trying actions other than the best one. We thus use the $\delta-$greedy algorithm before selection. More specifically, under a certain state, we select the best action according to the Q-table with a probability $1-\delta$ and randomly select one of the other actions with the probability $\delta$. Following the existing literature, at the beginning of the training, we set a relatively large $\delta$ so that we can try more; then, with the training process proceeds, we gradually reduce $\delta$ until the Q-table is converged and then Algorithm 1 is terminated. 

\begin{figure}[t]
	\begin{center}
		\includegraphics[width=.75\linewidth] {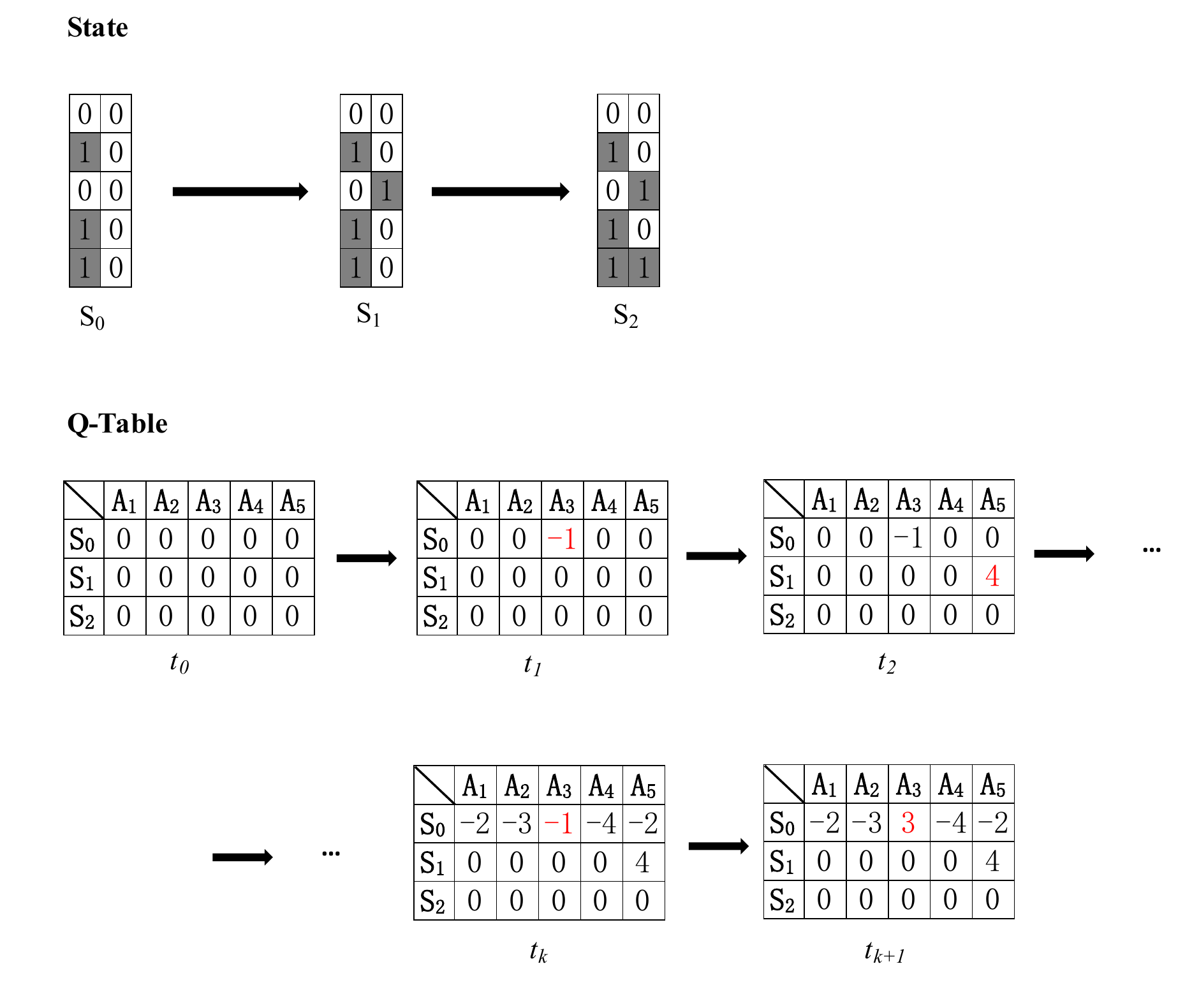}
		\vspace{-1.5em}
		\caption{An illustrative example of tabular Q-learning.}
		\label{fig:q_learning_example}
	\end{center}
\end{figure}

Figure~\ref{fig:q_learning_example} illustrates an example of using tabular Q-learning for training Q-function. For simplicity, we set the discount factor $\gamma$ to 1 and the learning rate $\alpha$ to 1. Here, we suppose that there are five cells in the target area, and we only consider two recent cycles: the last and current one. Hence, the state $\mathbf S$ has the dimension of $5\times 2$, as shown in $\mathbf{S}_0$, $\mathbf{S}_1$, and $\mathbf{S}_2$. The value $1$ means that the cell has been selected and $0$ means not. First, we initialize the table, $i.e.$ all the values in the Q-table are set to 0. When we first meet some states, e.g., $\mathbf{S}_0$, scores of all the actions in the Q-table under $\mathbf S_0$ are 0 (Q-table: $t_0$ in Figure~\ref{fig:q_learning_example}). We then randomly select one action since all the values are equal. If we choose the action $\mathbf{A}_3$ (select the cell 3), the state turns to $\mathbf{S}_1$. Then we update Q[$\mathbf S_0, \mathbf A_3$] as the current reward score plus the maximum score of the next state $\mathbf S_1$ (i.e., future reward). The current reward is $-c$ since the current cycle cannot satisfy the quality requirement ($c=1$ in the example). The maximum score for the state $\mathbf S_1$ is 0 in the Q-table. Hence, we get $Q[\mathbf{S}_0,\mathbf{A}_3]=-1+0=-1$ (Q-table: $t_1$ in Figure~\ref{fig:q_learning_example}). Similar, under $\mathbf{S}_1$, we choose $\mathbf{A}_5$. If these selections could satisfy the quality, we get the current reward is $R-c=4$ ($R$ is set to $5$, i.e., total number of cells). Also, the maximum possible reward of the next state $\mathbf S_2$ is 0 in the current Q-table. Then we update $Q[\mathbf{S}_1,\mathbf{A}_5]=5-1+0=4$ (Q-table: $t_2$ in Figure~\ref{fig:q_learning_example}). After some rounds, we meet $\mathbf{S}_0$ many times and maybe other actions are not good. And the Q-table would be changed to Q-table: $t_k$ in Figure~\ref{fig:q_learning_example}. This time, under $\mathbf{S}_0$, we check Q-table and find that $\mathbf{A}_3$ has the largest value, so we choose and perform $\mathbf{A}_3$. Then, we update $Q[\mathbf{S}_0,\mathbf{A}_3]=-1+4=3$, since the maximum reward score of the next state $\mathbf{S}_1$ is 4 (Q-table: $t_{k+1}$ in Figure~\ref{fig:q_learning_example}). Therefore, at the next times when we meet $\mathbf{S}_0$ again, we would probably choose the action $\mathbf{A}_3$, since it has the largest reward score.

While the tabular Q-learning can work well for an MCS task in a target area including a small number of cells, as shown in the above example, practical MCS tasks may involve a large number of cells. Suppose there are $50$ cells in the target area and we want to consider recent $2$ cycles to model states, then the state space will become extremely huge, $|\mathbb S|=2^{2\times 50}=2^{100}$, which is intractable in practice. To overcome this difficulty, in the next subsection we will propose to leverage deep learning with reinforcement learning to train the decision function for cell selection in Sparse MCS.

\subsection{Training Q-function with Deep Recurrent Q-Network}

To overcome the problem incurred by the extremely large state space in the cell selection, we then turn to use the Deep Q-Network (DQN), which combines Q-learning with deep neural networks. The difference between DQN and tabular Q-learning is that a deep neural network is used to replace the Q-table to deal with the dimension curse. In DQN, we do not need the Q-table lookups, but calculate $Q[\mathbf{S},\mathbf{A}]$ for each state-action pair selection. More specifically, the DQN inputs the current state and action, then it uses a deep neural network to obtain an estimated value of $Q[\mathbf{S},\mathbf{A}]$, shown as
{\begin{align}
Q(\mathbf{S},\mathbf{A})=\mathbb{E} \big[\mathbf{R}+\gamma max_{\mathbf{A}'} Q(\mathbf{S}',\mathbf{A}') \big]
\label{eq3}
\end{align}}%

For each selection, we use the neural network parameterized by $\theta$ to calculate the Q-function and select the best state-action pair who has the largest reward score, or called Q value. Note that the $\delta-$greedy algorithm is also used in DQN to balance the exploration and exploitation.

To obtain the estimation of the Q value which approximates the expected one in (\ref{eq3}), our proposed DQN uses the \textit{experience replay} technique. After one selection, we obtain the experience at current time step $t$, denoted as $\mathbf{e}_t=\langle \mathbf{S},\mathbf{A},\mathbf{R},\mathbf{S}'\rangle$, and the memory pool is $\mathbf{D}=\{\mathbf{e}_1,\mathbf{e}_2,...,\mathbf{e}_t\}$. Then, DQN randomly chooses part of the experiences to learn and update the network parameters $\theta$. The goal is to calculate the best $\theta$ to obtain $Q_{\theta}\approx Q$. The stochastic gradient algorithm is applied with the learning rate $\alpha$ and the loss function is defined as follow,
{\begin{align}
L(\theta_t)=\mathbb{E}_{\langle \mathbf{S},\mathbf{A},\mathbf{R},\mathbf{S}' \rangle} \Big[\big(\mathbf{R}+\gamma max_{\mathbf{A}'}Q_{\theta_t}(\mathbf{S}',\mathbf{A}')- 
Q_{\theta_t}(\mathbf{S},\mathbf{A})\big)^2 \Big]
\label{eq4}
\end{align}}%
Thus
{\begin{align}
\nabla_{\theta_t}L(\theta_t)=\mathbb{E}_{\langle \mathbf{S},\mathbf{A},\mathbf{R},\mathbf{S}' \rangle}
\Big[\big(\mathbf{R}+\gamma max_{\mathbf{A}'} Q_{\theta_t}(\mathbf{S}',\mathbf{A}')- Q_{\theta_t}(\mathbf{S},\mathbf{A})\big)\nabla_{\theta_t}Q_{\theta_t}(\mathbf{S},\mathbf{A})\Big]
\end{align}}%

For each update, DQN randomly chooses part of experiences from $\mathbf{D}$, then calculates and updates the network parameters $\theta$. Moreover, to avoid the oscillations (i.e., the Q-function changes too rapidly in training), we apply the \textit{fixed Q-targets technique}. More specifically, we do not always use the latest network parameter $\theta_t$ to calculate the maximum possible reward of the next state (i.e., $max_{\mathbf{A}'}Q_{\theta_t}(\mathbf{S}',\mathbf{A}')$), but update the corresponding parameter $\theta'$ every a few iterations, i.e.,
\begin{align}
	\nabla_{\theta_t}L(\theta_t)=\mathbb{E}_{\langle \mathbf{S},\mathbf{A},\mathbf{R},\mathbf{S}' \rangle}
	\Big[\big(\mathbf{R}+\gamma max_{\mathbf{A}'} Q_{\theta'}(\mathbf{S}',\mathbf{A}')- Q_{\theta_t}(\mathbf{S},\mathbf{A})\big)\nabla_{\theta_t}Q_{\theta_t}(\mathbf{S},\mathbf{A})\Big]
	\label{eq5}
\end{align}
The DQN learning algorithm is summarized in Algorithm~\ref{Alg2}.

\begin{algorithm}[t]
\renewcommand{\algorithmicrequire}{\textbf{Initialization:}}
\caption{Deep Recurrent Q-Network Learning}
\label{Alg2}
\begin{algorithmic}[1]
\REQUIRE ~~\\
    Initialize the network parameters, $t=0$

\WHILE{True}

    \STATE{Update the current state $\mathbf{S}=[\mathbf{s}_{-k},..., \mathbf{s}_{-1}, \mathbf{s}_0]$}
    \STATE{Calculate Q-value by Deep Q-Network with the parameter $\theta_t$ via (\ref{eq3}), select  action $\mathbf{A}$ with $\delta$-greedy algorithm.}

    \IF{The selected cells in current cycle satisfy the quality requirements}
        \STATE{// \emph{The cell selection in this cycle is complete, next state is the initial state in next cycle.}}
        \STATE{Update the next state $\mathbf{S}'=[\mathbf{s}_{-k+1},..., \mathbf{s}_{0}, \mathbf{s}_1]$}
        \STATE{Obtain the reward for this action $\mathbf{R}=R-c$}
    \ELSE
        \STATE{// \emph{Continue to select cells in this cycle.}}
        \STATE{Update the next state $\mathbf{S}'=[\mathbf{s}_{-k},..., \mathbf{s}_{-1}, \mathbf{s}_0']$}
        \STATE{Obtain the reward for this action $\mathbf{R}=-c$}
    \ENDIF
    \STATE{$\mathbf{e}_t=\langle \mathbf{S},\mathbf{A},\mathbf{R},\mathbf{S}'\rangle\ \rightarrow\ \mathbf{D}$}
    \STATE{Randomly select some $\mathbf{e}$ from $\mathbf{D}$}
    \STATE{Calculate $\theta_t$ via (\ref{eq5})}
    \STATE{$t++$}
    \IF{$t\% $RPLACE$\_$ITER$\ ==\ 0$}
        \STATE{ $\theta' = \theta_t$}
    \ENDIF
    
\ENDWHILE
\end{algorithmic}
\end{algorithm}

In DQN, how to design the network structure also impacts the effectiveness of the learned Q-function. One common way is using dense layers to connect the input (\textit{state}) and output (\textit{a reward score vector of all possible actions}). However, the temporal correlations exist in our state $\mathbf S$, but the dense layers cannot catch the temporal pattern well. We thus propose to use LSTM (Long-Short-Term-Memory) layers rather than dense layers in DQN so as to catch the temporal patterns in our state, which is also called \textit{Deep Recurrent Q-Network} (DRQN) \cite{DRQN}. More specifically, in DRQN, Q-function can be defined as,
\begin{equation}
	Q_{\theta_t}(\mathbf{O}_t,\mathbf{H}_{t-1},\mathbf{A})
\end{equation}
where  $\mathbf{O}_t$ represents the observation at time step $t$ (i.e., the cell selection vector at $t$), and $\mathbf{H}_{t-1}$ is the extra input returned by the LSTM network from the previous time step $t-1$. In our cell selection problem, a state $\mathbf{S}=[\mathbf{s}_{-k+1},..., \mathbf{s}_{-1}, \mathbf{s}_0]$ can be divided into $k$ time steps of observations, and then can also be used as inputs of the DRQN for learning the Q-function. 


\subsection{Reducing Training Data by Transfer Learning}

With deep reinforcement learning, we can get the Q-function that outputs reward scores for all the possible actions under a certain state, and then we can choose the cell that has the largest score in cell selection. However, the Q-function learning algorithm mentioned in the previous sections may need a large amount of training data, which also incurs collection cost for MCS organizers.\footnote{An organizer needs to conduct a preliminary study on the target sensing area to collect the data from every cell for a short time.} Then, can we reduce the amount of training data under certain circumstances?

In reality, many types of data have inter-data correlations, e.g., temperature and humidity \cite{Wang2017SPACE}. Then, if there are multiple correlated sensing tasks in a target area, probably the cell selection strategy learned for one task can benefit another task. With this intuition, we present a transfer learning method for learning the Q-function of an MCS task (\textit{target} task) with the help of the cell selection strategy learned from another correlated task (\textit{source} task). We assume that  the \textit{source} task has adequate training data, while the \textit{target} task has only a little training data. Inspired by the fine-tuning techniques widely used in image processing with deep neural networks, for training the Q-function of the target task, we initialize the parameters of its DRQN to the parameter values of the source task DRQN (learned from the adequate training data of the source task). Then, we use the limited amount of training data of the target task to continue the DRQN learning process (Algorithm~\ref{Alg2}). In such a way, we can reduce the amount of training data required for obtaining a good cell selection strategy of the target task.



\section{Evaluation}

In this section, we conduct extensive experiments based on two real-life datasets, which include various types of sensed data in representative MCS applications, such as temperature, humidity, and air quality. 

\subsection{Datasets}
Same as previous Sparse MCS literature \cite{Wang2017SPACE,wang2015ccs}, we adopt two real-life datasets, \emph{Sensor-Scope}~\cite{Fran2010SensorScope}, and \emph{U-Air}~\cite{Zheng2013U}, to evaluate the performance of our proposed cell selection algorithm \textit{DR-Cell}. These two datasets contain various types of sensed data, including temperature, humidity, and air quality. 
The detailed statistics of the two datasets are listed in Table~\ref{Table1}. Although these sensed data are collected from sensor networks or static stations, the mobile devices can also be used to obtain them (as in \cite{devarakonda2013real,hasenfratz2012participatory}). 
We can treat them as the data sensed by smartphones and use these datasets in our experiments to show the effectiveness of DR-Cell. 

\begin{table}[t]

      \centering
      \caption{Statistics of Two Evaluation Datasets}

      \begin{tabular}{|c|c|c|}
        \hline
        \multirow{2}{*}{} & \multicolumn{2}{c|}{\textbf{Datasets}} \\
         \cline{2-3}
        & \textbf{\emph{Sensor-Scope}} & \textbf{\emph{U-Air}}\\
        \hline
        City & Lausanne (Switzerland) &Beijing (China)\\
         \hline
        Data & temperature, humidity & PM2.5\\
         \hline
        Cell size ($m^2$) & 50*30 & 1000*1000 \\
         \hline
        Number of cells & 57 & 36 \\
         \hline
        Cycle length ($h$)& 0.5 & 1\\
         \hline
        Duration ($d$) & 7 & 11\\
         \hline
        Error metric & mean absolute error & classification error\\
         \hline
        \multirow{2}{*}{Mean $\pm$ Std.} & $6.04\ \pm\ 1.87\ ^\circ C$ (temperature) & \multirow{2}{*}{$79.11\ \pm \ 81.21$ (PM2.5)}\\
         \cline{2-2}
         & $84.52\ \pm\ 6.32\ \%$ (humidity) & \\
         \hline
      \end{tabular}
      \label{Table1}
\end{table}

\emph{Sensor-Scope}~\cite{Fran2010SensorScope}: The \emph{Sensor-Scope} dataset contains various environment readings, $e.g.$ temperature and humidity. The sensed data are collected from the EPFL campus with an area about $500m\times300m$. We first divide the target area into 100 cells, each cell is $50m\times30m$. Then we find that 57 out of 100 cells are deployed with valid sensors. Hence, we use the sensed data at the 57 cells to evaluate our algorithms. We use the mean absolute error to measure the inference error.

\emph{U-Air}~\cite{Zheng2013U}: The \textit{U-Air} dataset includes the air quality readings from Beijing. Same as \cite{Zheng2013U}, we split the Beijing to cells where each cell is $1km \times 1km$. Then, there are 36 cells with the sensed air quality readings. With this dataset, we conduct the experiment of PM2.5 sensing, and try to infer the air quality index \textit{category}\footnote{Six categories \cite{Zheng2013U}: Good (0-50), Moderate (51-100), Unhealthy for Sensitive Groups (101-150), Unhealthy (150-200), Very Unhealthy (201-300), and Hazardous (>300)} of unsensed cells. The inference error is measured by classification error.

\subsection{Baseline Algorithms}

We compare DR-Cell to two existing cell selection methods: QBC and RANDOM.

\textbf{QBC}: Based on the researches in active learning on matrix completion, Wang et al. \cite{Wang2017SPACE} present an intuitive method, called Query by Committee based cell selection algorithm. QBC selects the salient cell determined by "committee" to allocate the next task. More specifically, QBC attempts to use some different data inference algorithms, such as compressive sensing and K-Nearest Neighbors, to infer the full sensing matrix. Then, it allocates the next task to the cell with the largest variance among the inferred values of different algorithms.

\textbf{RANDOM}: In each sensing cycle, RANDOM will randomly select cells one by one until the selected cells can ensure a satisfying inference accuracy. 

\subsection{Experiment Process}

To learn DR-Cell, we use the first 2-day data of each dataset to train our Q-function, i.e., we suppose that the MCS organizers will conduct a 2-day preliminary study to collect data from all the cells of the target area. After the 2-day training stage, we enter the testing stage when we can use the trained Q-function to obtain the reward of every possible action under the current state, and then choose the action (i.e., cell) with the largest reward score. During the testing stage, we use the leave-one-out Bayesian inference method to ensure $(\epsilon, p)$-quality, same as previous Sparse MCS literature \cite{Wang2017SPACE}. The objective is to select cells as few as possible with the quality guarantee, and thus we will compare the number of cells selected by DR-Cell and baseline methods to verify the effectiveness of DR-Cell.

\subsection{Experiment Results}

\begin{figure}[t]
	\begin{center}
		\includegraphics[width=.75\linewidth] {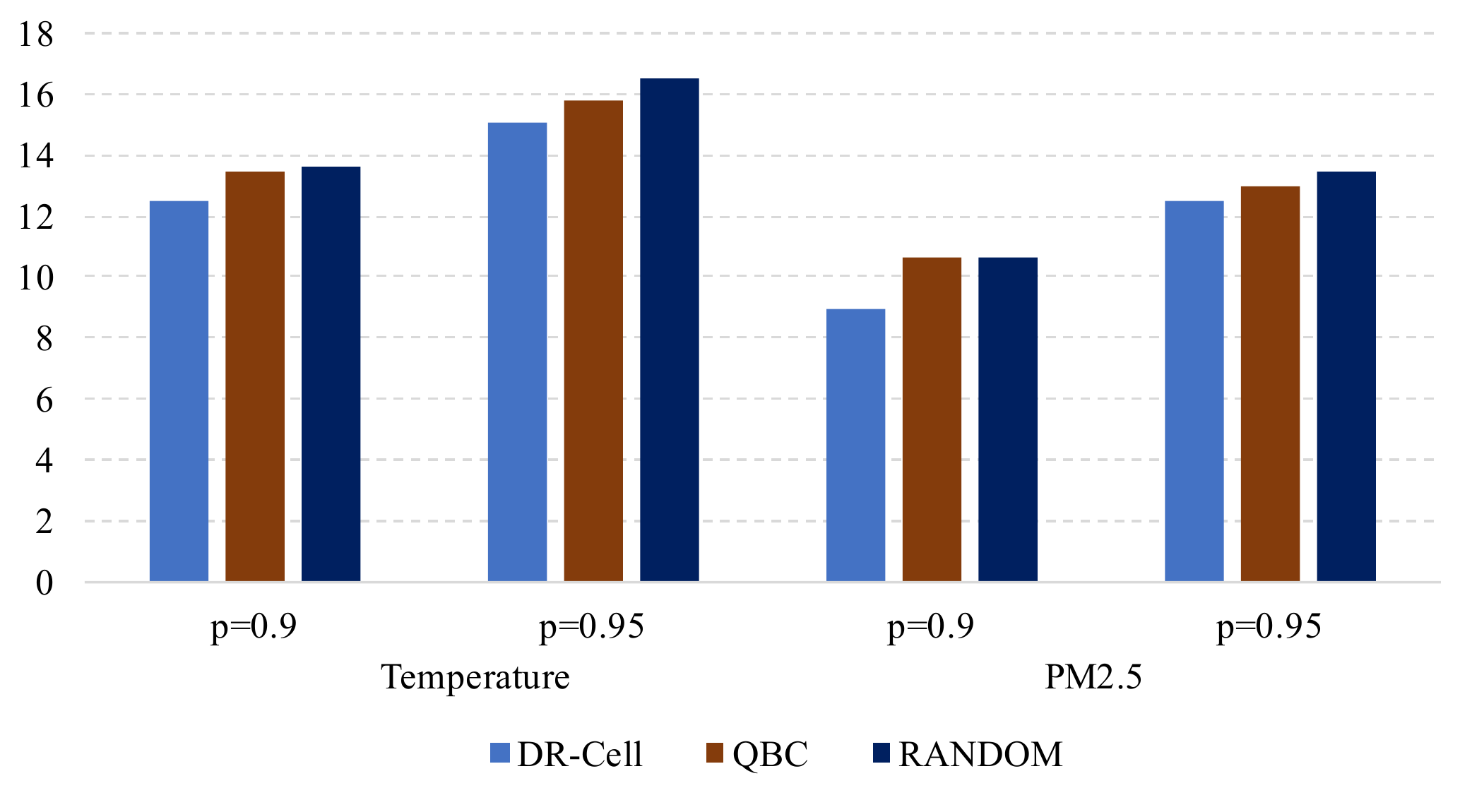}
		\vspace{-1em}
		\caption{Number of selected cells for temperature and PM2.5 sensing tasks.}
		\label{fig:eval1}
	\end{center}
\end{figure}

We first evaluate the performance by using the temperature data in \emph{Sensor-Scope} and the PM2.5 data in \emph{U-Air}, respectively. The results are shown in Figure~\ref{fig:eval1}.

In the temperature scenario of \emph{Sensor-Scope}, for the predefined ($\epsilon$-$p$)-quality, we set the error bound $\epsilon$ as $0.3^\circ C$ and $p$ as $0.9$ or $0.95$. This quality requirement is that the inference error is smaller than $0.3^\circ C$ for around 90\% or 95\% of cycles. 
Figure~\ref{fig:eval1} (leftmost part) shows the average numbers of selected cells for each sensing cycles. DR-Cell always outperforms two baseline methods. More specifically, when $p=0.9$, DR-Cell can select $6.9\%$ and $7.9\%$ fewer cells than QBC and RANDOM, respectively. 
In general, DR-Cell only needs to select 12.84 out of 57 cells for each sensing cycle when ensuring the inference error below $0.3^\circ C$ in $90\%$ of cycles.
When we improve the quality requirement to $p=0.95$, DR-Cell needs to select more cells to satisfy the higher requirement. Particularly, DR-Cell selects 15.08 out of 57 cells under the $(0.3^\circ C$-$0.95)$-quality and achieves better performances by selecting $4.6\%$ and $8.5\%$ fewer cells than QBC and RANDOM, respectively. 
For the PM2.5 scenario in \emph{U-Air}, we set the error bound $\epsilon$ as $9/36$ and $p$ as $0.9$ or $0.95$ and get the similar observations shown in Figure~\ref{fig:eval1} (rightmost part). When $p$ is $0.9$/$0.95$, DR-Cell selects 9.0/12.5 out of 36 cells and reduces $15.4\%$/$4.1\%$, and $15.5\%$/$7.3\%$ of selected cells than QBC and RANDOM, respectively. 


We then conduct the experiments on the multi-task MCS scenario, i.e., temperature-humidity monitoring, in \emph{Sensor-Scope} to verify the transfer learning performance. We conduct two-way experiments, $i.e.$ temperature as the source task and humidity as the target task; and vice versa. More specifically, for the source task, we still suppose that we obtain 2-day data for training; but for the target task, we suppose that we only obtain 10 cycles (i.e., 5 hours) of training data. 
Moreover, we add two compared methods to verify the effectiveness of our transfer learning method: \textbf{NO-TRANSFER} and \textbf{SHORT-TRAIN}. NO-TRANSFER is the method that directly uses the Q-function of the source task to the target task, and SHORT-TRAIN means that the target task model is only trained on the 10-cycle training data. 

\begin{figure}[t]
	\begin{center}
		\includegraphics[width=.75\linewidth] {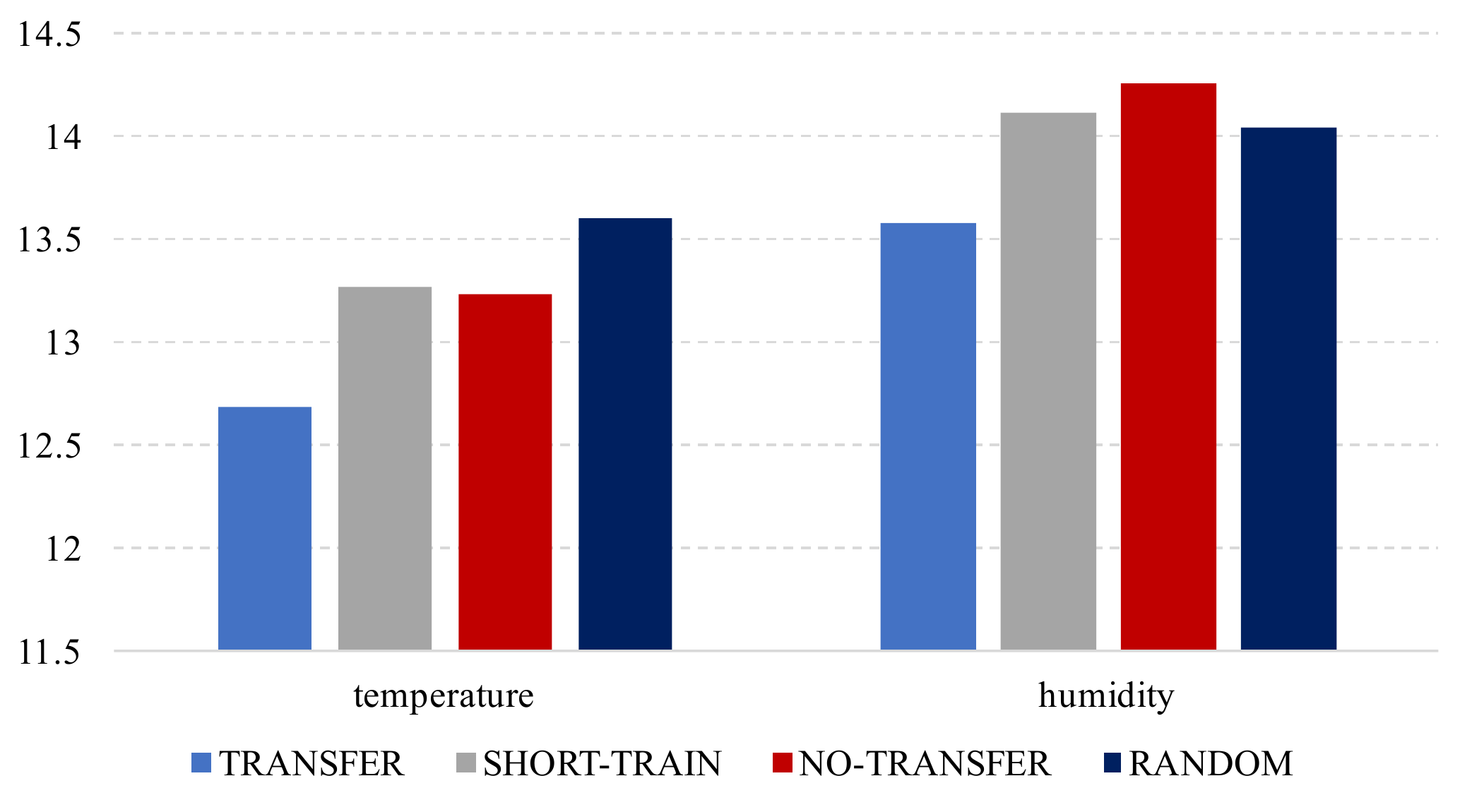}
		\vspace{-1em}
		\caption{Number of selected cells for temperature and humidity sensing tasks (transfer learning).}
		\label{fig:eval2}
	\end{center}
\end{figure}

The quality requirement of temperature is ($0.3^\circ C$-$0.9$)-quality and the humidity is $(1.5\%$-$0.9)$-quality. Figure~\ref{fig:eval2} shows the average numbers of selected cells. When temperature is seen as the target task, TRANSFER can achieve better performance by reducing $4.1\%$, $4.3\%$, and $6.7\%$ selected cells compared with NO-TRANSFER, SHORT-TRAIN, and RANDOM, respectively. When humidity is the target task, similarly, TRANSFER can select $3.8\%$, $4.7\%$, and $3.3\%$ fewer cells than NO-TRANSFER, SHORT-TRAIN, and RANDOM, respectively. Note that NO-TRANSFER and SHORT-TRAIN even perform worse than RANDOM in this case. It emphasizes the importance of having an adequate amount of training data for DR-Cell. By using transfer learning, we can significantly reduce the training data required for learning a good Q-function in DR-Cell, and thus further reducing the data collection costs of MCS organizers.

Finally, we report the computation time of DR-Cell. Our experiment platform is equipped with Intel Xeon CPU E2630 v4 @ 2.20GHz and 32 GB RAM. We implement our DR-Cell training algorithm in TensorFlow (CPU version). In our experiment scenarios, the training time consumes around 2--4 hours, which is totally acceptable in real-life deployments as the training is an off-line process.

\section{Conclusion}

In this paper, to improve the cell selection efficiency in Sparse MCS, we propose a novel \textit{Deep Reinforcement} learning based \textit{Cell} selection mechanism, namely \textit{DR-Cell}. We properly model the three key concepts in reinforcement learning, i.e., state, action, and reward, and then propose a deep recurrent Q-network with LSTM to learn the Q-function that can output the reward score given an arbitrary state-action pair. Then, under a certain state, we can choose the cell with the largest reward score as the next cell for sensing. Furthermore, we propose a transfer learning method to reduce the amount of training data required for learning the Q-function, if there are multiple correlated MCS tasks conducted in the same target area. Experiments on various real sensing datasets verify the effectiveness of DR-Cell in reducing the data collection costs. 

In our future work, we will study how to conduct the reinforcement learning based cell selection in an online manner, so that we do not need a preliminary study stage for collecting the training data any more. Besides, we will also consider a case where the data collection costs of different cells are diverse. Finally, we will consider to extend our mechanism to multi-task allocation scenarios when heterogeneous tasks are conducted simultaneously \cite{wang2017psallocator,wang2018multi} and privacy-preserving scenarios when the participant privacy protection mechanisms are applied \cite{wang2016differential,wang2017location,wang2017geographic}.

\bibliographystyle{ACM-Reference-Format}
\bibliography{rl-sparse-mcs}

\end{document}